\documentclass[preprint,12pt]{elsarticle}

\usepackage{amssymb}
\usepackage{amsmath}
\usepackage{multirow}

\DeclareMathOperator*{\argmin}{argmin}

\newtheorem{Definition}{Definition}

\journal{Computer Vision and Evolutionary Algorithms}

\begin{document}

\begin{frontmatter}



\title{Analytical-Heuristic Modeling and Optimization for Low-Light Image Enhancement}


\author[label1]{Axel Martinez}   
\author[label1]{Emilio Hernandez} 
\author[label2]{Matthieu Olague}
\author[label1]{Gustavo Olague\corref{cor1}} 
\cortext[cor1]{Corresponding author}
\affiliation[label1]{organization={Ensenada Center for Scientific Research and Higher Education},
            addressline={Km 107 Carretera Tijuana-Ensenada}, 
            city={Ensenada},
            postcode={22860}, 
            state={Baja California},
            country={Mexico}}

\affiliation[label2]{organization={IBM Technology Campus Guadalajara},
	addressline={El Salto}, 
	city={Guadalajara},
	postcode={45680}, 
	state={Jalisco},
	country={Mexico}}

\begin{abstract}
Low-light image enhancement remains an open problem, and the new wave of artificial intelligence is at the center of this problem. This work describes the use of genetic algorithms for optimizing analytical models that can improve the visualization of images with poor light. Genetic algorithms are part of metaheuristic approaches, which proved helpful in solving challenging optimization tasks. We propose two analytical methods combined with optimization reasoning to approach a solution to the physical and computational aspects of transforming dark images into visible ones. The experiments demonstrate that the proposed approach ranks at the top among 26 state-of-the-art algorithms in the LOL benchmark. The results show evidence that a simple genetic algorithm combined with analytical reasoning can defeat the current mainstream in a challenging computer vision task through controlled experiments and objective comparisons. This work opens interesting new research avenues for the swarm and evolutionary computation community and others interested in analytical and heuristic reasoning.
\end{abstract}

\begin{graphicalabstract}
	\centering
\includegraphics[scale=.25]{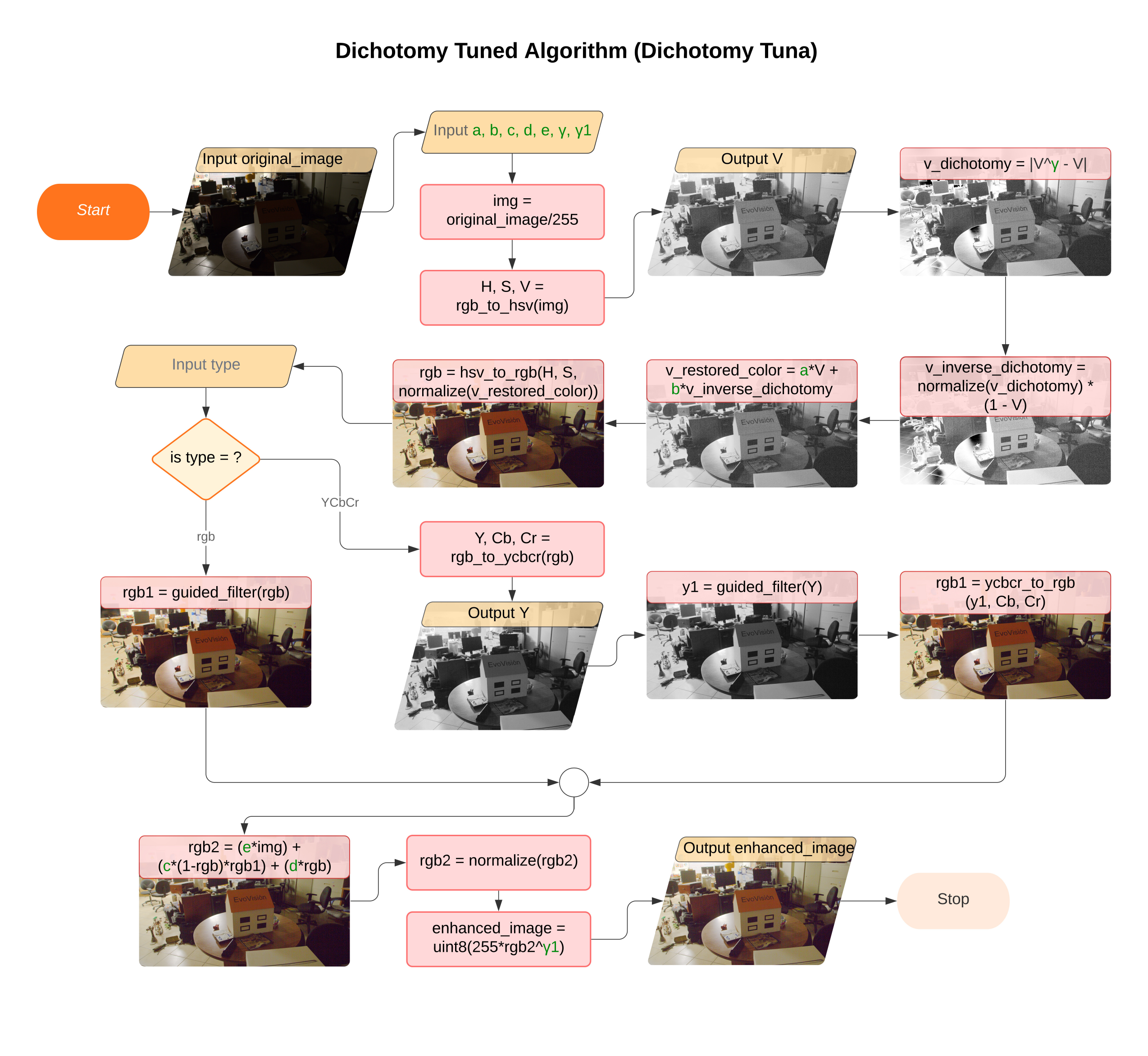}
\end{graphicalabstract}

\begin{highlights}
\item This article shows how classical meta-heuristics can be at the forefront of low-light image enhancement.
\item Genetic algorithms combined with analytical-heuristic reasoning led to a high-quality method for low-light image enhancement.
\item The combination of analytical reasoning and optimization is at the top compared to 26 state-of-the-art methods. 
\item The proposed analysis-synthesis approach gives insight into low-light image enhancement.
\item The method, characterized by its simplicity and clarity, follows attributes that modern artificial intelligence demands.
\end{highlights}


\end{frontmatter}

\section{Introduction}

An analytical model represents a formal analysis to address specific questions in science and engineering. Analytical design aims to explain concepts and methods to understand or solve challenging tasks or endeavors precisely and transparently. Mathematics is the language of analytical models, and the main goal is to provide quantitative or computational models or systems through equations by specifying parametric relationships with their associated parameter values as a function of time, space, and other fundamental variables. On the other hand, metaheuristics are high-level procedures designed to discover, generate, or select specific parameter values or lower-level procedures that may provide a sufficiently good solution, usually within an optimization framework. A golden rule non-written among people studying metaheuristics is that the problem should be highly complex to apply such methods. In other words, if there is an analytical solution, metaheuristics should not be used. Therefore, part of the traditional problem statement is to demonstrate that the subject of interest needs an analytical solution. Nevertheless, the analytical models are often too complicated, and heuristics are necessary to approach a solution. A balanced view of analysis and synthesis is usually required to achieve the best possible solution by combining these two approaches. We hold to this position as exposed in \cite{Olague2016}, where the study of computer vision follows the evolutionary computation paradigm, proposing a balanced approach between analysis and synthesis.

This article presents a method developed from the viewpoint of physics and applied mathematics in combination with heuristic search mechanisms that shed light on a challenging computer vision problem. Low-light image enhancement is a complex computer vision problem, and several surveys testify to the increasing interest of several communities in solving this problem \cite{Li2022, Guo2023, Ye2024}. Nevertheless, metaheuristic algorithms are not part of the state-of-the-art field of study as portrayed in the surveys. Moreover, images with low-light issues prevent developers from effectively accomplishing designs in many fields, such as medical images, automatic driving, surveillance, object detection, and a long list. Developing a practical approach to this problem through analytical-heuristic reasoning can attract people working with low-light image enhancement to study metaheuristic approaches in artificial intelligence tasks.

Nowadays, machine learning through deep learning attracts considerable interest, and the topic of low light is no exception. However, approaching the problem through a purely analytical-synthetic methodology should be interesting if the method shows its value in a conventional benchmark. The reward of having such a method could open new avenues from the viewpoint of accuracy, reliability, efficiency, and power consumption. To arrive at those goals, we must first design an analytical optimization method that shows significant value in pursuing a path away from the mainstream.

We organize this article in four different sections. First, we survey the best methods published at main venues to identify the approaches that could be of value when making a fair comparison. Second, we introduce an analytical model and its extension with heuristic reasoning, combining both with a well-known metaheuristic (genetic algorithm) that makes the foundation of our approach to the problem. Third, we provide experimental results using a widely applied benchmark that could serve as a standard for all methods surveyed in this article. Fourth, we finalize the article by discussing the opportunities that bring our method and a conclusion highlighting the benefits and challenges while providing a list of future research avenues in low-light image enhancement.

\section{Related Work}

Low-light image enhancement improves the perception or interpretability of images taken under poor illumination. Technical and artistic issues arise due to several factors generally reflected in poor contrast, such as the scene composition or environment where the photographer is taking the picture or simply due to the physical constraints of the camera in what is known as the exposure triangle. Recently, people from several communities have started to look closer at the problem, mainly due to the flourishing of the deep learning paradigm based on neural networks. In the following, we review some classical and deep learning methods that could represent current state-of-the-art techniques. The selected works include a rich collection of approaches ranging from conventional methods to convolutional neural networks (CNNs), attention guided U-Net, multiscale networks, generative adversarial networks, and transformer-based learning. This section aims to recount the preferred methods used to compare our proposed methodology through a benchmark that all methods use to establish a different approach that people from other communities not included in the mainstream could use as inspiration for their work. This includes communities devoted to metaheuristics, mathematical and physical modeling, and optimization. Moreover, the proposed technique could be blended with the mainstream, as we will appreciate from studying the following methods.

\subsection{Classical Low-light Image Enhancement Methods}

First, we review what can be classified as classical methods for low-light image enhancement.
The first method is Adaptive Gamma Correction with Weighting Distribution (AGCWD) \cite{ref-journal5}. The method centers on modifying histograms while enhancing contrast in digital images. The work provides an automatic transformation technique to improve the brightness of dimmed images via the gamma correction and probability function of luminance pixels. This method is representative of classical solutions to the problem of low-light image enhancement that does not require learning and, therefore, can be used in many situations as a ready-to-use technique. The following method is known as Natural Preserved Enhancement (NPE) \cite{ref-journal14}. This enhancement algorithm follows the retinex theory, which aims to improve the image from the viewpoint of lightness and naturalness. The solution is a bi-log transformation that maps illumination while balancing details and naturalness. Next, Simultaneous Reflectance and Illumination Estimation (SRIE) \cite{ref-journal15}. This method follows a variational approach of retinex modeling, where physical image modeling results from the product of reflectance and illumination. The model considers gamma correction for illumination adjustment with noise suppression. Another method is the Fusion-based Enhancing Method (FBEM) \cite{ref-journal16}. This article presents a fusion-based framework for enhancing weakly illuminated images using a single image and a simple illumination estimation algorithm based on morphological closing. The fusion stage can adopt different techniques, resulting in a plurality of results while being susceptible to GPU implementation. A highly cited work on our list is known as Low light image enhancement (LIME) \cite{ref-journal17}. This work follows an augmented Lagrangian multiplier approach based on the retinex theory that refines an initial illumination map by imposing a structure along the R, G, and B channels, resulting in better aesthetically pleasing images. Like other similar approaches, this method is computationally intensive and may require specialized hardware. The following selected method is the Semi-decoupled decomposition (SDD) \cite{ref-journal20}. This method follows the retinex theory by gradually decomposing the information on the illumination layer via a Gaussian total variation model while considering the reflectance layer and noise in the analysis. Among several metrics, the authors selected the Natural Image Quality Evaluator (NIQE) to assess performance, a non-reference metric that is a widely accepted criterion.

\subsection{Deep-learning Methods for Low-light Image Enhancement}

We now provide a select set of deep-learning approaches to low-light image enhancement. The first method is the Deep Retinex Network (Retinex-Net) \cite{ref-journal18}, which enhances the image in a three-step decomposition, adjustment, and reconstruction process. First, Retinex-Net decomposes the image into illumination and reflectance; then, it brightens up the illumination, and finally, concatenates illumination and reflectance following a multi-scale perspective on illumination while removing noise along reflectance as a final step. This work introduces the LOL database, which provides low-light and normal pairs of images that researchers later use as a benchmark with several metrics. The second method is the Learning to restore images via decomposition and enhancement(LRIDE) \cite{ref-journal19}. This work focus on the study of low-light images contaminated with noise and the fact that frequency correlates with noise. The idea is to detect objects in the low-frequency and then enhancing high-frequency details. Authors use a set of images with noise called SID (learning to see in the dark) and two popular metrics PSNR (Peak Signal-to-Noise Ratio) and SSIM (Structural Similarity Index Metric) with good results \cite{ref-journal35}. Another technique is the Deep local parametric filters (DeepLPF) \cite{ref-journal21}. The approach applies three kinds of learned spatially local filters: elliptical, graduated, and polynomial. The authors designed a neural network that regresses the parameters of those filters that are, in turn, applied to enhance the image. The authors compare their approach to images obtained from the MIT-Adobe-5K-DPE and MIT-Adobe-5K-UPE databases using the PSNR and SSIM metrics where human artists retouched images. We can appreciate that the database selection correlates with the research goals. The method Zero-Reference Deep Curve Estimation (Zero-DCE) formulates light enhancement as a task of image-specific curve estimation \cite{ref-journalZdce}. The model considers a lightweight deep network to estimate pixel-wise and high-order curves for dynamic range adjustment while considering value range, monotonicity, and differentiability. The method does not require pair or unpaired data during training since it uses a set of carefully formulated non-reference loss functions. Nevertheless, the final results include PSNR and SSIM metrics on GPU technology.

The following algorithm in our selection is the Pre-trained Image Processing Transformer (IPT) \cite{ref-journal22}, representing a new trend in deep learning characterized by its outstanding results. This technology bases its approach on the increasing computer power and large-scale datasets. The authors trained the transformer on the well-known ImageNet benchmark and proposed that anyone can apply the pre-trained model to different tasks after fine-tuning the model. PSNR is a metric that authors suggested to compare the results fairly for high-resolution tasks. Another technique is the Multi-scale residual dense network (MS-RDN) \cite{ref-journal23}. This work recognizes the wide range of aspects devoted to low-light image enhancement, like noise, contrast, and color bias. This work follows the retinex theory while designing an end-to-end network with layer-specified constraints and data-driven mapping for single-image low-light enhancement. The authors proposed a sparse gradient minimization sub-network to preserve edge information while removing low-amplitude structures. Then, two new networks focus on illumination and reflectance enhancement. The authors test the algorithm with the LOL dataset and several metrics, such as PSNR and SSIM. 

The EnlightenGan (EG) model is a generative adversarial network that creates normal and low image spaces without exactly paired images \cite{ref-journal24}. The article considers deep learning better than traditional methods using low-normal image pairs. However, the work overlooks the fact that the deep learning community considers an open problem learning with few images. The aim is to create larger datasets through the generative approach and evaluate them with the no-referenced metric NIQE. The Retinex-inspired unrolling with architecture search (RUAS) is a lightweight architecture based on retinex theory \cite{ref-journal25}. Authors recognize deep learning increasing over-engineering architecture design and high computational burden. However, users pose questions from the project page since they cannot replicate the results. The authors adopt PSNR and SSIM as part of their metrics, along with the LOL database and other image pair data. The authors use a powerful graphics card, TITAN X GPU, to conduct the experiments. 

Another method is the U-shaped transformer (Uformer) \cite{ref-journal26}. In this work, authors deploy the transformer-based architecture to leverage the capability of self-attention maps at multi-scale resolution to recover more image details. The authors evaluate the proposal with PSNR and SSIM on several datasets. This work is an adaptation of a transformer architecture to low-light image enhancement. The self-calibrated illumination (SCI) recognizes that existing techniques are unperformed in unknown complex scenarios \cite{ref-journal27}. The aim is to create a framework for fast, flexible, and robust learning in low-light scenarios. Authors use PSNR and SSIM reference metrics and NIQE and LOE (lightness order error) no-reference metrics on several datasets. The solution exhibits an overengineering design that requires GPU cards. The article describing the Signal-to-Noise-Ratio (SNR) method explores transformers and convolutional networks to enhance pixels with spatial-varying operations dynamically \cite{ref-journal28}. The article focuses on noise, leaving aside traditional aspects devoted to manipulating color, tone, and contrast. The method develops around the SNR ratio to improve noise and visibility. The experiments consider the LOL database, among others, using PSNR and SSIM metrics. Adaptivity and efficiency are suitable characteristics pursued in the Retinex-based deep unfolding network (URetinex-Net) \cite{ref-journal29}. Like classical methods, the idea is to apply regularization under the retinex scheme. The aim is to improve the design from the viewpoints of data dependency, high efficiency in the search process, and user-specified illumination enhancement. The benefits are noise suppression and detail preservation. The authors provide results on the LOL database, including various methods evaluated with PSNR and SSIM metrics.

The following method in our list is the Cell Vibration Model (CVM) \cite{ref-journal30}. The idea turns around energy modeling and gamma correction to develop new mathematical modeling for global lightness enhancement. Authors recognize that current approaches are prone to over-enhancement, color distortion, or time consumption. The new relationship between image lightness and gamma intensity provides the base for a local fusion strategy. Results contrast the proposed solution against several methods on several databases, including LOL under PSNR, NIQE, and BRISQUE (Blind/Referenceless Image Spatial Quality Evaluator) metrics.
Another technique is the Deep Recursive Band Network (DRBN), which considers linear band representation with the guidance of paired low and normal light images. However, training with unpaired data is possible \cite{ref-journal31}. The results considered the PSNR and SSIM metrics, with the LOL dataset showing visually pleasing contrast, color distributions, and well-restored structural details. The Kindling the Darkness (KinD) is another retinex-inspired network that uses paired images shot under different exposition conditions \cite{ref-journal32}. The method runs on GPU technology while applying classical optimization. Authors use the LOL dataset, among others, to compare popular low-light image enhancement methods with the PSNR, SSIM, LOE, and NIQE metrics. Authors alleviate visual defects left in KinD with a multi-scale illumination attention module while considering similar metrics and datasets \cite{ref-journalknd++}. 

Finally, we include four deep networks from 2023 to complete the list. An unsupervised method based on retinex theory (PairLIE) adaptively learns priors from low-light image pairs \cite{ref-journal33}. The method considers extremely dark cases of the LOL dataset and measures it with PSNR and SSIM metrics. Another method proposes a one-stage retinex-based framework with an illumination-guided transformer called Retinexformer \cite{ref-journal35}. The algorithm proposes to first estimate the illumination information to light up the low-light image and then restore the corruption to enhance the image. Authors test the proposal with the complete LOL dataset, among others, using the PSNR and SSIM metrics. In another work authors apply the transformer technology in a method called LLFormer using attention blocks to reduce complexity \cite{ref-journal36}. Despite introducing a new ultra-high-definition benchmark authors include results with the LOL dataset while continuing to use the PSNR and SSIM metrics. The approach is computationally intensive and pre-trained models are available. The Illumination-aware Gamma Correction (IAGC) is a deep learning method that incorporates gamma correction to improve the quality of results \cite{ref-journal37}. The network follows the trend of applying a transformer block to create a local-to-global hierarchical attention mechanism. The results are exclusively develop under the LOL dataset with the PSNR and SSIM metrics. The method requires GPU technology with numerous parameters and extensive learning.

In summary, the reviewed methods are highly cited and known for the robustness and quality of their results. The most recent methods follow the deep learning paradigm with overly designed techniques that turn them obscure due to the black-box nature of deep neural networks. To alleviate this issue, many methods use classical approaches like gamma correction, retinex theory, and variational and Lagrange optimization. We can appreciate that since the first deep learning methods; the LOL database has been part of the experimental section on most papers, together with reference metrics (PSNR and SSIM) and non-reference metrics (NIQE, BRISQUE, and LOE). The issue of attaining real-time performance is a problem, and some authors focus on reducing the complexity of solutions.

\section{Methodology}

This section outlines the methodology, composed of two main parts: an analytical model and an optimization method. We provide two models based on the mathematical reasoning developed in \cite{unpublished2024}. Then, we optimize the parameters with the classical genetic algorithm \cite{inbook2}. The low-light image enhancement problem considers an image $I \in \mathbb{R}^{W \times H \times 3}$ of width $W$ and height $H$ in the RGB color space. The following relationship can represent the process of image enhancement:

\begin{equation}\label{function}
	\hat{R} = f(I; \theta) \; ,
\end{equation}

\noindent
where $\hat{R} \in \mathbb{R}^{W \times H \times 3}$ is the enhanced result, and $f$ represents a function with parameters $\theta$. The purpose of mathematical modeling is to find optimal model parameters $\hat{\theta}$ that minimize the error
	
\begin{equation}\label{minimization}
	\hat{\theta} = \argmin_{\theta} Q(\hat{R},R) \; ,
\end{equation}

\noindent
where $R \in \mathbb{R}^{W \times H \times 3}$ is the ground truth, and the quality function $Q(\hat{R},R)$ drives the model optimization. The idea is then to propose a model and an optimization framework.

\subsection{Mathematical Model}

This section introduces the proposed model based on gamma correction, which is usually applied to adjust the luminance or tristimulus values in video or still image systems. 

\begin{Definition}{\bfseries[Gamma correction]} \label{gamma_correction}
	Given an input image $I$ with values $I(x,y,z)$ and the transformed image $I'$ with output values $\hat{R}(x,y,z)$. Usually, the relationship follows a power law.
	
	\begin{equation}
		\hat{R}(x,y,z) = I(x,y,z)^\gamma \;\;\; 0 \leq I \leq 1 , \gamma \in \mathbb{R}^+\;\;\;.
	\end{equation}
	
	\noindent
	The image increases the contrast in dark areas and loses it in light areas when $\gamma < 1$, while for $\gamma > 1$, the image increases its contrast in light areas and decreases it in dark places.
\end{Definition}

This mathematical model has difficulty handling the dynamic range to adequately express the issue of tonal contrast due to its characteristic of monotonically increasing in the first quadrant. The following relationship provides the mathematical model applied in our work:

\begin{Definition}{\bfseries[Dichotomy function]} \label{dichotomy}
	The model considers an input image $I$ with values $I(x,y,z)$ and the transformed image $I'$ with output values $\hat{R}(x,y,z)$. The following relationship describes a behavior where the output is strictly monotonically increasing in the first part until the output reaches a maximum, and the second part is characterized by its output being strictly monotonically decreasing.
	\begin{equation} \label{dichotomy_equation}
		\hat{R}=k|I(x,y,z)^{\gamma}-I(x,y,z)| \;,
	\end{equation}
	
	\noindent
	where the term $I(x,y,z) = I(x,y,z)^{1}$ represents the identity, while $I(x,y,z)^{\gamma}$ with $\gamma \in \{\mathbb{R}^+ \cup 0\}$ is the correction applied to the input value. The difference between the contrasted function and the reference function $\hat{R}=k|I(x,y,z)^{\gamma}-I(x,y,z)|$ produces a function with a dichotomous behavior that magnifies the contrast produced by the transformation function, where $k$ is a scale factor to normalize the output between $0 \leq \hat{R}(x,y,z) \leq 1$. 
\end{Definition}

\noindent
The resulting equation maximizes the difference of contrasts $DoC$ with two main slopes (positive and negative) and a unique inflection point where the maximum contrast located at $d_{\max}$ is $DoC_{\max}= \max(I(x,y,z)^{\gamma_-1})$.

\subsection{Dichotomy Tuned Algorithm}
\label{Tuna}

The proposed dichotomy function greatly enhances low-light images that contain dark regions in the whole picture. Thus, the original dichotomy function attempts to improve image information extraction for different computer vision tasks \cite{unpublished2024}. On the other hand, the low-light image enhancement problem requires that the output image contains visually pleasant information. There is an issue in this way since the dichotomy function enhances visual information without considering aesthetics. In the following, we propose a design based on this relationship that considers uneven, poorly illuminated areas where high-lights contrast with low-lights in some parts.

\begin{figure}[ht]
	\centering
	\includegraphics[scale=0.33]{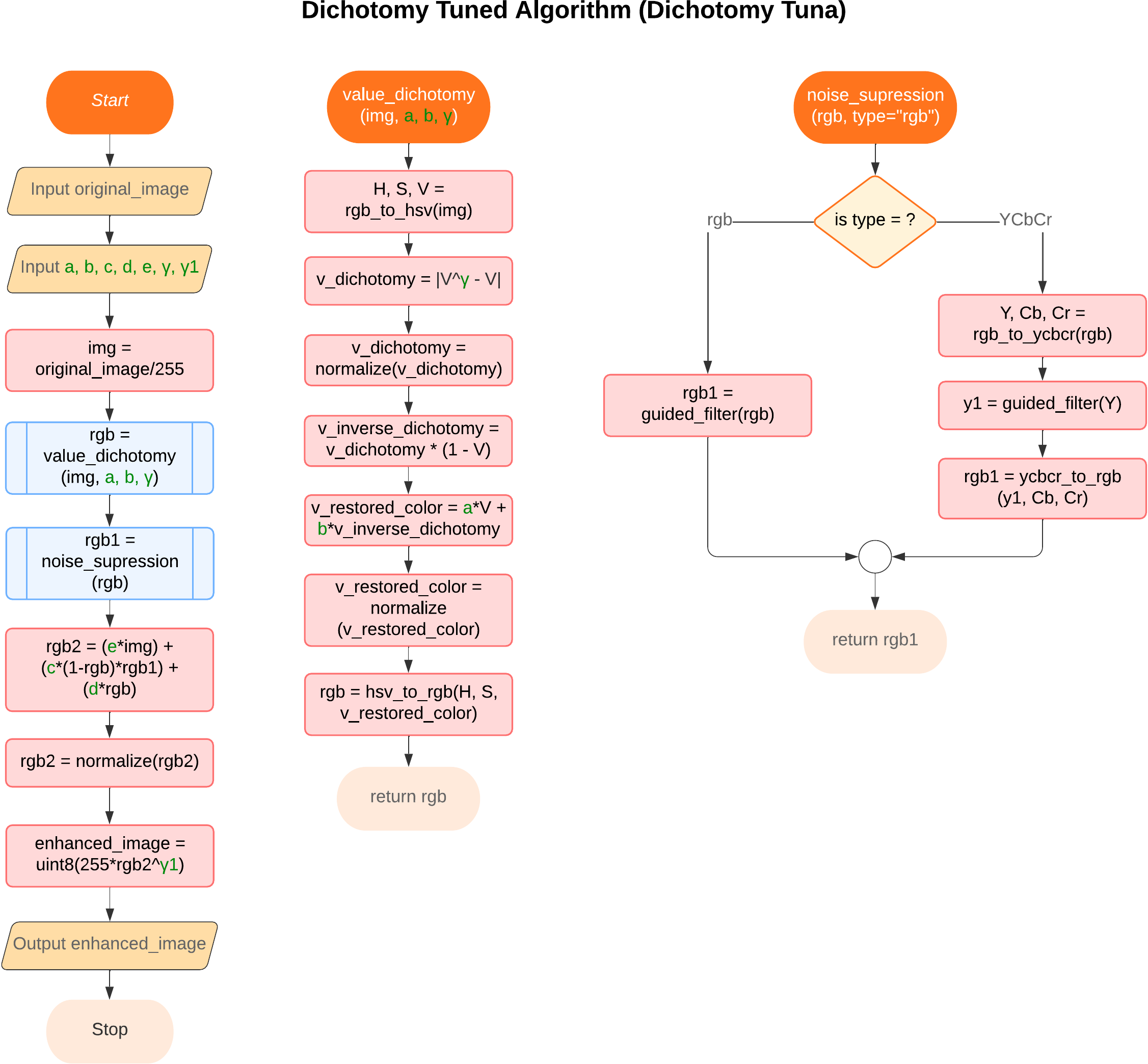}
	\caption{This flowchart shows all the steps we consider to improve the original Dichotomy model to highlight the visual appearance of images with under and over-exposed light regions. 
	\label{fig-Tuna}}
\end{figure}  

We propose a new method, coined Dichotomy Tuned Algorithm (Dichotomy Tuna), which receives as inputs an image $I \in \mathbb{R}^{W \times H \times 3}$ of width $W$ and height $H$ in the RGB color space, and seven parameters $(a,b,c,d,e.\gamma, \gamma1)$. Figure \ref{fig-Tuna} shows all the necessary steps that we describe next.

The original image is first normalized on a scale from 0 to 1 by dividing the pixel values by 255. The resulting output image $img$ is then passed to a method that leverages the dichotomy function by converting $img$ from the RGB to the HSV color space, extracting the values $H, S$, and $V$. We then applied the dichotomy function to V by using the value of parameter $\gamma$, whose output goes through a normalization process with the min-max function:

\begin{equation}
	x'=\frac{x-\min(x)}{\max(x) -\min(x)} \; .
\end{equation}

\noindent
The result $v_{dichotomy}$ then undergoes a color restoration phase consisting of two stages. First, we multiply $v_{dichotomy}$ by the inverse of $V$ to prevent color inversion in the second stage of the color restoration. Then, $v_{restore\_color}$ results from multiplying the original value $V$ by a parameter $a$ and adding it to the result of multiplying $v_{inverse\_dichotomy}$ by a parameter $b$. The result of this operation, $v_{\_restored\_color}$, is normalized once again and returned to the RGB color space by using the original $H$ and $S$ values. The image obtained from this process undergoes a noise suppression stage, which by default applies a guided filter to the RGB image itself, resulting in a filtered image $rgb1$. However, a less computationally expensive filtering option, which converts the RGB image to the YCbCr color space, is also available. In this option, we propose using a guided filter to the $Y$ channel of the YCbCr image and apply the result $y1$ when converting back to the RGB color space, yielding an alternative $rgb1$ image. In all reported experiments, we followed the first approach.

The last stage of Dichotomy Tuna leverages computed images $img, rgb, rgb1$ and parameters $(c,d,e)$ to compute the resulting image $rgb2$ through the following relationship:

\begin{equation}
	rgb2 = (e * img) + (c * (1-rgb) * rgb1) + (d * rgb) \; .
\end{equation}

\noindent
Finally, $rgb2$ is normalized and processed through a final gamma correction stage by leveraging the value $\gamma1$. The output is then multiplied by 255 and converted to an integer of type uint8 for each channel, which results in the final enhanced image.

\subsection{Optimization with Genetic Algorithm}

This section deals with the optimization problem of finding the best parameters to improve contrast enhancement for low-light images. We decided to apply the well-known genetic algorithm, which many considered outdated, mainly because we want to highlight the importance of the proposed models based on the Definition (\ref{dichotomy}). We pose the optimization problem regarding maximization since we selected a fitness (objective) function as a linear combination of two popular metrics (PSNR and SSIM) whose values increase, corresponding to better results.

\begin{equation}
	Q(\hat{R},R) = \max\{\rho(PSNR) + \tau(SSIM)\} \; ,
\end{equation}

\noindent
where $\rho = 1/100$ and $\tau = 1$. Note that obtaining the maximum of $Q()$ and its corresponding maximizing or maximal policy, unique or not, corresponds to a minimization problem, except that it is necessary to reverse all involved functions according to the simplest reasoning.

The next step in formulating a solution to the above problem is to represent a possible solution with a chromosome $\Gamma = \{\gamma_{1\ldots n}\}$ for Equation (\ref{dichotomy_equation}), encoded as a vector of real numbers of the same length. Since a well-taken photograph requires specific parameters and not a model containing all possible parameters, we decided to find the best parameter set for each picture in the database. We use a population of size 10 for each parameter.

The proposed genetic algorithm recombines each parameter with the SBX crossover operator \cite{ref-journal34}. This operator emulates the working principles of single-point crossover on binary strings. The method considers two parent solutions $P_1$ and $P_2$, and creates two children $C_1$ and $C_2$ as follows:

\begin{eqnarray*}
	C_1 & = & 0.5 [ (1+\beta)P_1 + (1-\beta)P_2 ], \nonumber \\
	C_2 & = & 0.5 [ (1-\beta)P_1 + (1+\beta)P_2 ], \nonumber 
\end{eqnarray*}

\begin{equation*}
	\text{with } \beta = \left \{ \begin{array}{l} (2u)^{\frac{1}{\eta_x+1}} \quad \;\;\;\; \text{ if } u < 0.5, \\ 
		(\frac{1}{2(1-u)})^{\frac{1}{\eta_x-1}} \;\;\; \text{ otherwise.} \end{array} \right.
\end{equation*}

The spread factor $\beta$ depends on a random variable $u \in [0,1]$ on a user-defined non-negative value $\eta_x=2$ that fixes the distribution of children to the parents. We apply mutation to each real variable of individuals using a polynomial distribution perturbation \cite{Olague2016}. This operator modifies a parent $P$ into a child $C$ using the boundary values $P^{(LOW)}$ and $P^{(UP)}$ of each decision variable as follows:

\begin{eqnarray*}
	C & = & P + \delta (P^{(UP)} - P^{(LOW)}) , \\
\end{eqnarray*}

\begin{equation*}
	\text{with } \delta = \left \{ \begin{array}{l} (2u)^{\frac{1}{\eta_m+1}} - 1 \quad \;\;\;\; \quad \text{ if } u < 0.5, \\ 
		1 -[2(1-u)]^{\frac{1}{\eta_m+1}} \;\;\; \text{ otherwise,} \end{array} \right.
\end{equation*}

\noindent
where $\delta$ is the spread factor dependent on a random variable $u \in [0,1]$ and a user defined non-negative value $\eta_m=25$ that characterizes the distribution of the resulting variable. We executed 50 algorithm runs during 50 generations each and selected the best solution for each dataset.

\section{Experimental Results}

\begin{figure}[h]
	\centering
	\includegraphics[width=14.0cm]{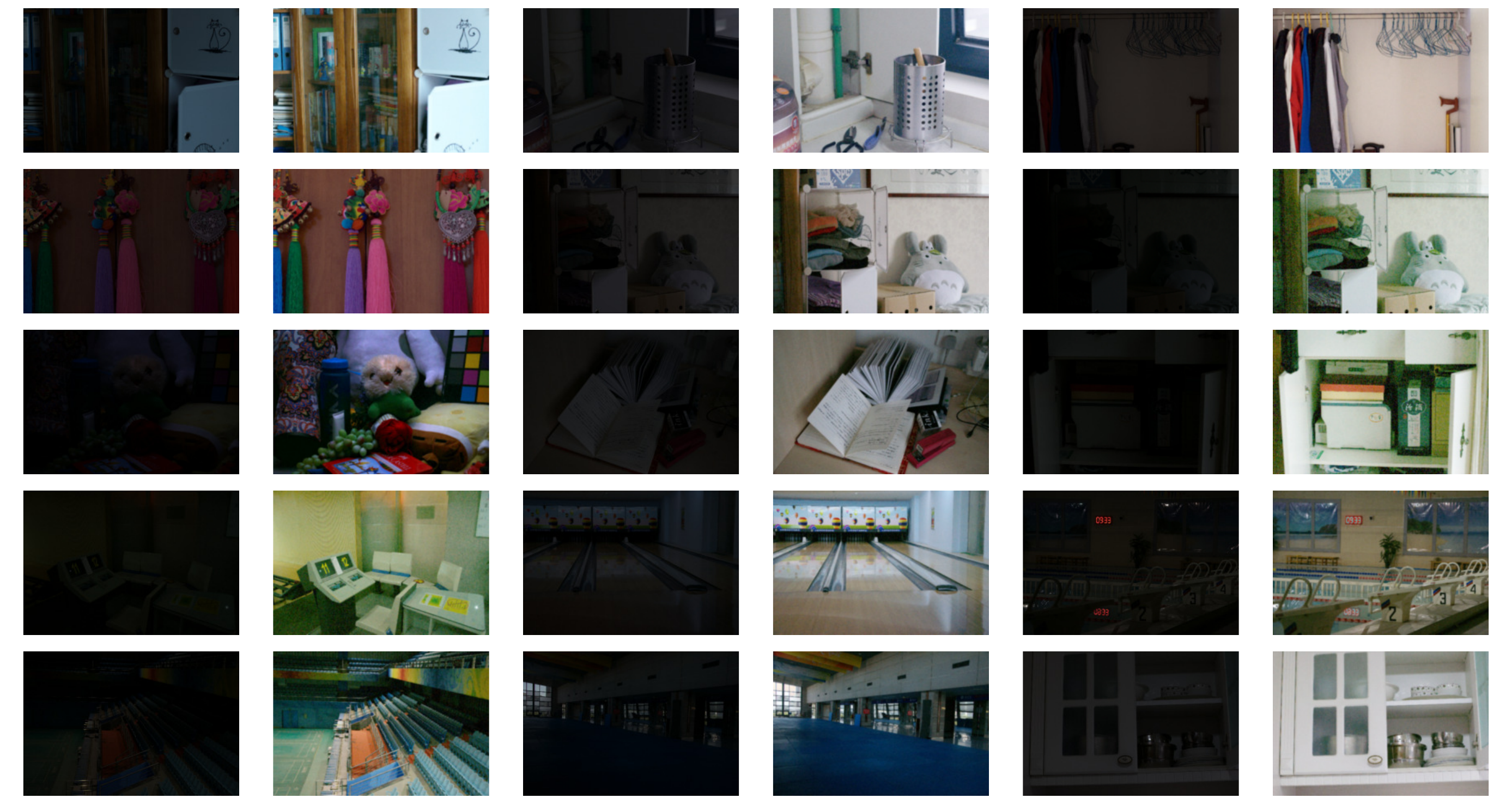}
	\caption{This collage shows the fifteen low-light testing images in the LOLv1 database, and the photos improved with the dichotomy model plus a Gaussian filter of $\sigma = 1$ presented in this work. Surprisingly, it reproduces tones like daytime, with a simple mathematical operation that contrasts state-of-the-art models based on complex reasoning. \label{fig4}}
\end{figure}

 We evaluate our method on the widely used LOL dataset, including LOLv1 and LOLv2. LOLv1 contains 485 real low-normal light pairs for training and 15 for testing \cite{ref-journal18}. LOLv1 was further improved, creating LOLv2, which includes two sets \cite{ref-journal23}. LOLv2-real consists of 689 pairs for training and 100 pairs for testing. LOLv2-real follows the exact characteristics of LOLv1, collecting low-normal light pair photographs by changing exposure time and ISO while fixing other camera configurations. The authors created LOLv2-synthetic considering the illumination distribution of synthetic images matching the property of real dark pictures. LOLv2-synthetic contains 900 synthetic low-normal light pairs for training and 100 for testing. Our method did not require learning and only used the testing images. Figure \ref{fig4} provides the result of our method after genetic algorithm optimization to the images of LOLv1. We use the peak signal-to-noise ratio (PSNR) and the measure of structural similarity (SSIM) as quantitative metrics \cite{ref-journalquality}. 
 
 \begin{table}[t]
 	\centering
 	\resizebox{\textwidth}{!}{%
 		\begin{tabular}{l c c c c c c c c}
 			\hline
 			Methods & AGCWD \cite{ref-journal5} & NPE \cite{ref-journal14} &  SRIE \cite{ref-journal15} & FBEM \cite{ref-journal16} & LIME \cite{ref-journal17} & Retinex-Net \cite{ref-journal18} & LRIDE \cite{ref-journal19} & SDD \cite{ref-journal20}\\ 
 			&(IEEE TIP 13)&(IEEE TIP 13)&(CVPR 16)&(SP 16)&(IEEE TIP 17)& (BMVC 18)&(CVPR 19)& (IEEE TM 20) \\
 			\hline
 			\hline
 			PSNR ↑ & 13.0462 ± 3.4439 & 16.96 & 11.86 & 16.96 & 16.76 & 16.7740 ± 2.6345 & 18.27 & 13.34\\
 			SSIM \ ↑ & 0.4047 ± 0.112859 & 0.481 & 0.493 & 0.505 & 0.444 & 0.5362 ± 0.0892 & 0.665 & 0.635\\
 			\hline	
 			Methods & DeepLPF \cite{ref-journal21} & Zero-DCE \cite{ref-journalZdce} & IPT \cite{ref-journal22} &  MS-RDN \cite{ref-journal23} & EG \cite{ref-journal24} & RUAS \cite{ref-journal25} & Uformer \cite{ref-journal26} & SCI \cite{ref-journal27} \\ 
 			&(CVPR 21)&(CVPR 20)&(CVPR 21)&(IEEE TIP 21)&(IEEE TIP 21)& (CVPR 21)&(CVPR 22)& (CVPR 22)\\
 			\hline
 			\hline
 			PSNR ↑ & 15.28 & 14.8607 ± 4.2719 &  16.27 & 17.20 & 17.48 & 16.4047 ± 4.3879 & 16.36 & 15.80 \\
 			SSIM \ ↑ & 0.473 & 0.6644 ± 0.1279 & 0.504 & 0.64 & 0.65 & 0.7036 ± 0.1379 & 0.507 & 0.527 \\
 			\hline
 			Methods   & SNR \cite{ref-journal28} & URetinex-Net \cite{ref-journal29} & CVM \cite{ref-journal30} &  DRBN \cite{ref-journal31} & KinD \cite{ref-journal32} & KinD++ \cite{ref-journalknd++} & PairLIE \cite{ref-journal33} & Retinexformer \cite{ref-journal35} \\ 
 			&(CVPR 22)&(CVPR 22) &(IEEE TM 22)&(IEEE TIP 21)&(ACM ICM 19)&(IJCV 21)&(CVPR 23) &(ICCV 23) \\
 			\hline
 			\hline
 			PSNR ↑ & 24.49 & 21.32 & 16.33 & 20.13 & 17.6476 ± 3.3405 & 17.7518 ± 2.7563 & 18.4684 ± 3.9020 & {\bfseries 25.1532 ± 2.7742} \\
 			SSIM \ ↑ & 0.840 & 0.835 & 0.583 & 0.830 & 0.8291 ± 0.0985 & 0.8163 ± 0.0794 & 0.8109 ± 0.0631 & {\bfseries 0.8985 ± 0.0408}\\
 			\hline
 			Methods & LLformer \cite{ref-journal36} & IAGC \cite{ref-journal37} & Dichotomy & Dichotomy & Dichotomy & & &  \\ 
 			& (AAAI CAI 23) & (CVPR 23) & & + filter & Tuna & & &\\
 			\hline
 			\hline
 			PSNR ↑ & 23.6491 ± 4.4786 & 24.53 & 23.0196 ± 2.6126 & {\bfseries 24.5866 ± 2.3471} & {\bfseries 23.9426 ± 2.1771 } & & &\\
 			SSIM \ ↑ & 0.8734 ± 0.0668 & 0.842 & 0.7277 ± 0.0851 & {\bfseries 0.8533 ± 0.055027} & {\bfseries 0.8063 ± 0.0586} & & & \\
 			\hline	
 		\end{tabular}
 	}
 	\caption{This table shows quantitative comparisons with state-of-the-art methods for image enhancement on the LOLv1 dataset using two standard metrics.}\label{Table1}
 \end{table}

 Table \ref{Table1} shows that our method obtains consistent best values by large margins against most methods. Our method achieves second place on PSNR and third place on SSIM. Note that these numbers are obtained either by running their respective codes or from their respective papers. In general, researchers report the average result without considering other statistical variables. We include averages with up to four significant figures and the standard deviation to help the comparison. We used original codes following the recommendations; however, some results did not match those reported in the original articles, so we kept the program results. Not all articles have code available; others require significant computer resources like high-costly graphics cards.
 
 \begin{table}[t]
 	\centering
 	\resizebox{\textwidth}{!}{%
 		\begin{tabular}{l c c c c c c c c}
 			\hline
 			Methods & AGCWD \cite{ref-journal5} & NPE \cite{ref-journal14} &  SRIE \cite{ref-journal15} & FBEM \cite{ref-journal16} & LIME \cite{ref-journal17} & Retinex-Net \cite{ref-journal18} & LRIDE \cite{ref-journal19} & SDD \cite{ref-journal20}\\ 
 			&(IEEE TIP 13)&(IEEE TIP 13)&(CVPR 16)&(SP 16)&(IEEE TIP 17)& (BMVC 18)&(CVPR 19)& (IEEE TM 20) \\
 			\hline
 			\hline
 			PSNR ↑ & 14.7287 ± 2.8891 & 17.33 & 17.34 & 18.73 & 15.24 & 16.0972 ± 1.6572 & 16.85 & 16.64\\
 			SSIM \ ↑ & 0.4482 ± 0.141641 & 0.461 & 0.686 & 0.559 & 0.470 & 0.5123 ± 0.1098 & 0.678 & 0.677\\
 			\hline	
 			Methods & DeepLPF \cite{ref-journal21} & Zero-DCE \cite{ref-journalZdce} & IPT \cite{ref-journal22} &  MS-RDN \cite{ref-journal23} & EG \cite{ref-journal24} & RUAS \cite{ref-journal25} & Uformer \cite{ref-journal26} & SCI \cite{ref-journal27} \\ 
 			&(CVPR 21)&(CVPR 20)&(CVPR 21)&(IEEE TIP 21)&(IEEE TIP 21)& (CVPR 21)&(CVPR 22)& (CVPR 22)\\
 			\hline
 			\hline
 			PSNR ↑ & 14.10 & 18.0587 ± 4.9268 &  19.80 & 20.06 & 18.23 & 15.3255 ± 3.4168 & 16.36 & 17.79 \\
 			SSIM \ ↑ & 0.480 & 0.6768 ± 0.1504 & 0.813 & 0.816 & 0.617 & 0.6727 ± 0.1262 & 0.771 & 0.568 \\
 			\hline
 			Methods   & SNR \cite{ref-journal28} & URetinex-Net \cite{ref-journal29} & CVM \cite{ref-journal30} &  DRBN \cite{ref-journal31} & KinD \cite{ref-journal32} & KinD++ \cite{ref-journalknd++} & PairLIE \cite{ref-journal33} & Retinexformer \cite{ref-journal35} \\ 
 			&(CVPR 22)&(CVPR 22) &(IEEE TM 22)&(IEEE TIP 21)&(ACM ICM 19)&(IJCV 21)&(CVPR 23) &(ICCV 23) \\
 			\hline
 			\hline
 			PSNR ↑ & 21.36 & 21.22 & 19.75 & 20.29 & 20.5882 ± 3.7322 & 17.6603 ± 3.1209 & 19.8845 ± 3.4771 & 22.7940 ± 3.7074 \\
 			SSIM \ ↑ & 0.842 & 0.860 & 0.627 & 0.831 & 0.8404 ± 0.0904 & 0.7913 ± 0.0869 & 0.8088 ± 0.0528 & 0.8666 ± 0.0478 \\
 			\hline
 			Methods & LLformer \cite{ref-journal36} & IAGC \cite{ref-journal37} & Dichotomy & Dichotomy & Dichotomy & & &  \\ 
 			& (AAAI CAI 23) & (CVPR 23) & & + filter & Tuna & & &\\
 			\hline
 			\hline
 			PSNR ↑ & {\bfseries 27.7488 ± 3.0447} & 22.20 & 22.5981 ± 2.8125 & {\bfseries 23.5804 ± 2.3582} & {\bfseries 24.1754 ± 2.9969} & & &\\
 			SSIM \ ↑ & {\bfseries 0.8881 ± 0.0402} & 0.863 & 0.7092 ± 0.1239 & {\bfseries 0.8232 ± 0.066277} & {\bfseries 0.8127 ± 0.0747 } & & & \\
 			\hline	
 		\end{tabular}
 	}
 	\caption{This table shows quantitative comparisons with state-of-the-art methods for image enhancement on the LOLv2-real dataset using two standard metrics.}\label{Table2}
 \end{table}

 Table \ref{Table2} provides the results of 29 methods, including our three variations of the dichotomy function. Again, one of our methods (Dichotomy Tuna) achieved second place in PSNR and fourth place in SSIM. In Table \ref{Table1}, Retinexformer achieves first place, while LLformer is the winner in this test. This last approach loses ten points on PSNR according to Table \ref{Table3}. It is hard to unveil why LLFormer has such a significant loss, which contrasts with our approach, which was easy to understand and fix, as we will observe in the following test. Besides, we use an outdated strategy to look for the best score, which contrasts with both deep learning methods being highly complex transformers and our approach's simplicity.

\begin{table}[t]
	\centering
	\resizebox{\textwidth}{!}{%
	\begin{tabular}{l c c c c c c c c}
		\hline
		Methods & AGCWD \cite{ref-journal5} & NPE \cite{ref-journal14} &  SRIE \cite{ref-journal15} & FBEM \cite{ref-journal16} & LIME \cite{ref-journal17} & Retinex-Net \cite{ref-journal18} & LRIDE \cite{ref-journal19} & SDD \cite{ref-journal20}\\ 
		&(IEEE TIP 13)&(IEEE TIP 13)&(CVPR 16)&(SP 16)&(IEEE TIP 17)& (BMVC 18)&(CVPR 19)& (IEEE TM 20) \\
		\hline
		\hline
		PSNR ↑ & 13.9131 ± 3.7947 & 16.59 & 14.50 & 17.50 & 16.88 & 17.1365 ± 1.9394 & 15.20 & 16.46\\
		SSIM \  ↑ & 0.6084 ± 0.181279 & 0.778 & 0.616 & 0.751 & 0.776 & 0.7937 ± 0.0716 & 0.612 & 0.728\\
		\hline	
		Methods & DeepLPF \cite{ref-journal21} & Zero-DCE \cite{ref-journalZdce} & IPT \cite{ref-journal22} &  MS-RDN \cite{ref-journal23} & EG \cite{ref-journal24} & RUAS \cite{ref-journal25} & Uformer \cite{ref-journal26} & SCI \cite{ref-journal27} \\ 
		&(CVPR 21)&(CVPR 20)&(CVPR 21)&(IEEE TIP 21)&(IEEE TIP 21)& (CVPR 21)&(CVPR 22)& (CVPR 22)\\
		\hline
		\hline
		PSNR ↑ & 16.02 & 17.7564 ± 3.8688 &  18.30 & 22.05 & 16.57 & 13.4041 ± 3.4825 & 19.66 & 17.73 \\
		SSIM \  ↑ & 0.587 & 0.8399 ± 0.1208 & 0.811 & 0.905 & 0.734 & 0.6764 ± 0.1189 & 0.871 & 0.763 \\
		\hline
		Methods   & SNR \cite{ref-journal28} & URetinex-Net \cite{ref-journal29} & CVM \cite{ref-journal30} &  DRBN \cite{ref-journal31} & KinD \cite{ref-journal32} & KinD++ \cite{ref-journalknd++} & PairLIE \cite{ref-journal33} & Retinexformer \cite{ref-journal35} \\ 
		&(CVPR 22)&(CVPR 22) &(IEEE TM 22)&(IEEE TIP 21)&(ACM ICM 19)&(IJCV 21)&(CVPR 23) &(ICCV 23) \\
		\hline
		\hline
		PSNR ↑ & 23.84 & 18.75 & 18.56 & 23.22 & 17.2758 ± 3.8237 & 17.4774 ± 3.0228 & 19.0744 ± 3.3130 & {\bfseries 25.6693 ± 4.9426} \\
		SSIM \  ↑ & 0.905 & 0.829 & 0.842 & 0.927 & 0.7808 ± 0.1495 & 0.8156 ± 0.0925 & 0.8439 ± 0.0655 & {\bfseries 0.9531 ± 0.0308} \\
		\hline
		Methods & LLformer \cite{ref-journal36} & IAGC \cite{ref-journal37} & Dichotomy & Dichotomy & Dichotomy & & &  \\ 
		& (AAAI CAI 23) & (CVPR 23) & & + filter & Tuna & & &\\
		\hline
		\hline
		PSNR ↑ & 17.1633 ± 3.2792 & 25.58 & 14.8082 ± 3.8282 & 15.0513 ± 3.9304 & {\bfseries 27.1717 ± 3.5416} & & &\\
		SSIM \  ↑ & 0.8275 ± 0.0906 & 0.940 & 0.7975 ± 0.1046 & 0.7328 ± 0.098561 & {\bfseries 0.9435 ± 0.0395} & & & \\
		\hline	
	\end{tabular}
}
	\caption{This table shows quantitative comparisons with state-of-the-art methods for image enhancement on the LOLv2-synthetic dataset using two standard metrics.}\label{Table3}
\end{table}

Table \ref{Table3} provides the final results on the LOLv2-synthetic dataset. We achieved first place in this test, with Retinexformer being the runner-up using PSNR. Surprisingly, LLformer drops ten points, unveiling a similar problem we encountered in our previous work. The dichotomy model drops around 8 points according to PSNR in Table \ref{Table2}, which was the primary motivation for developing a variant presented in Section \ref{Tuna}. Nevertheless, our approach follows analytical reasoning, and we could discern why the model failed. The poor result was due to a problem of reversing the output when there is an overexposition. We make a heuristic proposal with seven parameters while incorporating filtering operations that help in the final results. Contrary to the simplicity of our methodology, deep learning represents an abstruse method that is hard to fix, and usually, authors use brute force to improve the processes, whether by increasing the database or the overall computational structure. Regarding SSIM, our method scores second place well above the others.

\begin{table}[]
	\resizebox{\textwidth}{!}{%
		\begin{tabular}{lcccc}
			\hline
			Method & \multicolumn{1}{l}{BRISQUE ↓} & \multicolumn{1}{l}{NIQE ↓} & \multicolumn{1}{l}{PIQUE ↓} & \multicolumn{1}{l}{LOE ↓} \\
			\hline
			(IEEE TIP 13) AGCWD \cite{ref-journal5}  & 28.4206 ± 7.4975    & 7.8563 ± 0.98481     & 39.8425 ± 7.0938      & {\bfseries 0.13259 ± 0.15274}         \\
			(BMVC 18) Retinex-Net \cite{ref-journal18}       & 39.586 ± 3.43       & 9.7296 ± 1.3807      & 57.6731 ± 6.0335      & 993.2925 ± 264.9362       \\
			(CVPR 20) Zero-DCE \cite{ref-journalZdce}          & 30.3051 ± 4.7218    & 8.2232 ± 1.1685      & 35.5059 ± 7.7622      & 209.4268 ± 112.4915       \\
			(CVPR 21) RUAS \cite{ref-journal25}              & 25.6931 ± 4.2267    & 5.9273 ± 1.0043      & {\bfseries 18.5084 ± 6.1344}   & 264.676 ± 129.3697        \\
			(ACM ICM 19) KinD \cite{ref-journal32}              & 26.6453 ± 9.8619    & 3.894 ± 0.81299      & 45.2404 ± 14.4752     & 492.9934 ± 244.8023       \\
			(IJCV 21) KinD++ \cite{ref-journalknd++}            & {\bfseries 25.0881 ± 11.0998}  & 4.0088 ± 0.96185  & 46.9154 ± 16.2167          & 966.7364 ± 441.9105   \\
			(CVPR 23) PairLIE \cite{ref-journal33}           & 28.9668 ± 6.4293    & 4.0376 ± 0.48395     & 26.2112 ± 6.0648      & 283.6094 ± 73.114         \\
			(CVPR 23) Retinexformer \cite{ref-journal35}     & 25.6313 ± 8.0627    & {\bfseries 2.9713 ± 0.56474}  & 28.1563 ± 7.6907     & 274.6837 ± 92.7358        \\
			(CVPR 23) LLformer \cite{ref-journal36}          & 25.7695 ± 6.9976    & 3.1579 ± 0.48374     & 22.1614 ± 5.0695      & 305.3189 ± 107.8528       \\
			Dichotomy         & 31.4174 ± 4.9362    & 8.2737 ± 1.1195      & 39.836 ± 8.3387       & {\bfseries 3.0789 ± 9.756}            \\
			Dichotomy+filter  & 37.1962 ± 4.0146    & {\bfseries 4.1633 ± 0.32753}  & 49.9499 ± 8.7785     & 303.6071 ± 100.797        \\
			\multicolumn{1}{l}{Dichotomy Tuna} & {\bfseries 27.8135 ± 10.7769}      & 5.5463 ± 0.89548     & {\bfseries 25.8753 ± 9.1477}            & 299.4073 ± 117.7318   \\
			\hline   
		\end{tabular}%
	}
	\caption{This table provides the final results of selected methods for LOLv1 considering the no-reference metrics.}
	\label{Table4}
\end{table}

In the following, we selected nine methods to compete against our three proposals using the no-referenced metrics BRISQUE, NIQE, PIQUE, and LOE. Table \ref{Table4} shows different best methods according to the metrics. Remarkably, one of the oldest and classical methods scored highest in LOE, with Dichotomy ranking second. Experts consider LOE a specially designed illumination metric since it represents the lightness order error reflecting an enhanced image's naturalness. On the other hand, all other metrics are far from assessing the essence of measuring accurate visual perception of humans \cite{Li2022}. Generally, our approach provides reasonable results in all three other evaluation methods, ranking in the middle of the final results. Table \ref{Table5} provides similar behavior with different winners. Again, the Dichotomy model scores second in LOE, the proposed Dichotomy+filter achieves sixth in NIQE, and Dichotomy tuna gets eighth in BRISQUE and fourth in PIQUE. Finally, Table \ref{Table6} shows a similar trend of having different best methods according to the no-referenced metrics. Dichotomy drops to the bottom in LOE, while Dichotomy Tuna ranks third in PIQUE, NIQE, and BRISQUE. We can observe that it is difficult for an algorithm to score high in all four metrics. However, the classical approach of AGWCD achieves reasonable results in all four metrics. Note that our proposed approach is susceptible to applying a minimization process considering these four metrics.

\begin{table}[]
	\resizebox{\textwidth}{!}{%
		\begin{tabular}{lcccc}
			\hline
			Method & \multicolumn{1}{l}{BRISQUE ↓} & \multicolumn{1}{l}{NIQE ↓} & \multicolumn{1}{l}{PIQUE ↓} & \multicolumn{1}{l}{LOE ↓} \\
			\hline
			(IEEE TIP 13) AGCWD \cite{ref-journal5}              & 30.1636 ± 7.199  & 8.3884 ± 1.2519  & 40.8412 ± 7.845                    & {\bfseries 0.21566 ± 0.2978}    \\
			(BMVC 18) Retinex-Net \cite{ref-journal18}        & 42.4772 ± 3.7671 & 10.563 ± 1.7057  & \multicolumn{1}{l}{61.267 ± 5.585} & 857.2549 ± 237.2675 \\
			(CVPR 20) Zero-DCE \cite{ref-journalZdce}           & 31.9651 ± 6.2877 & 8.7667 ± 1.5677  & 41.0567 ± 8.4932                   & 204.5563 ± 89.7794  \\
			(CVPR 21) RUAS \cite{ref-journal25}               & 25.2394 ± 7.1376 & 6.1722 ± 1.275   & {\bfseries 17.1487 ± 5.8401}       & 320.0258 ± 148.2416 \\
			(ACM ICM 19)KinD \cite{ref-journal32}               & 27.5127 ± 9.925  & 4.1406 ± 1.0372  & 44.6449 ± 12.7555                  & 450.9918 ± 127.6265 \\
			(IJCV 21) KinD++ \cite{ref-journalknd++}             & 29.5345 ± 8.5301 & 4.2009 ± 1.0866  & 50.6639 ± 14.4463                  & 870.0179 ± 424.614  \\
			(CVPR 23) PairLIE \cite{ref-journal33}            & 28.4996 ± 9.0564 & 4.5369 ± 0.8642  & 25.9524 ± 6.9958                   & 351.8825 ± 127.5538 \\
			(CVPR 23) Retinexformer \cite{ref-journal35}      & {\bfseries 23.2907 ± 7.824}  & 3.5964 ± 0.78206 & 20.6822 ± 5.8527       & 352.0665 ± 140.7445 \\
			(CVPR 23) LLformer \cite{ref-journal36}           & 25.2512 ± 7.6904 & {\bfseries 3.4882 ± 0.57639} & 18.666 ± 5.9879        & 311.8023 ± 109.0256 \\
			Dichotomy          & 33.5866 ± 5.9206 & 8.9788 ± 1.6064  & 45.1834 ± 9.5407                   & {\bfseries 2.8262 ± 9.2872}     \\
			Dichotomy+filter & 39.2747 ± 3.8093 & {\bfseries 4.8046 ± 0.72707} & 43.9178 ± 9.3494         & 424.0145 ± 158.8758 \\
			\multicolumn{1}{l}{Dichotomy Tuna} & {\bfseries 30.3209 ± 11.9008}   & 5.8882 ± 1.1609        & {\bfseries 24.5327 ± 8.1014}            & 48.9413 ± 38.3726  \\
			\hline      
		\end{tabular}%
	}
	\caption{This table provides the final results of selected methods for LOLv2-real considering the no-reference metrics.}
	\label{Table5}
\end{table}

\begin{table}[]
	\resizebox{\textwidth}{!}{%
		\begin{tabular}{lcccc}
			\hline
			Method & \multicolumn{1}{l}{BRISQUE ↓} & \multicolumn{1}{l}{NIQE ↓} & \multicolumn{1}{l}{PIQUE ↓} & \multicolumn{1}{l}{LOE ↓} \\
			\hline
			(IEEE TIP 13) AGCWD \cite{ref-journal5}                   & 24.7915 ± 8.7422  & 4.2778 ± 1.3357 & 42.7112 ± 10.8951 & {\bfseries 11.415 ± 14.3285}    \\
			(BMVC 18) Retinex-Net \cite{ref-journal18}             & 28.3861 ± 10.0807 & 5.6904 ± 1.5964 & 44.262 ± 8.3491   & 500.5159 ± 276.803  \\
			(CVPR 20)Zero-DCE \cite{ref-journalZdce}                & 26.3645 ± 9.1709  & 4.357 ± 1.1716  & 40.4682 ± 10.1249 & 131.3644 ± 81.63    \\
			(CVPR 21) RUAS \cite{ref-journal25}                    & 37.4517 ± 9.4813  & 5.0917 ± 1.6208 & 50.4714 ± 14.2423 & 711.9155 ± 430.1154 \\
			(ACM ICM 19)KinD \cite{ref-journal32}                    & 30.1328 ± 9.3147  & 4.2548 ± 1.2369 & 49.9125 ± 11.9428 & 207.9232 ± 88.8515  \\
			(IJCV 21) KinD++ \cite{ref-journalknd++}                  & 29.0519 ± 9.0877  & 4.7659 ± 2.245  & 50.897 ± 10.8387  & 374.7168 ± 221.3175 \\
			(CVPR 23) PairLIE \cite{ref-journal33}                 & 32.2622 ± 9.3184  & 4.9457 ± 4.0279 & 47.1186 ± 12.1123 & 230.6041 ± 143.4428 \\
			(CVPR 23) Retinexformer \cite{ref-journal35}           & {\bfseries 24.4991 ± 8.7354}  & {\bfseries 3.9378 ± 1.0827} & 34.0766 ± 8.0624  & 161.5423 ± 69.7081  \\
			(CVPR 23) LLformer \cite{ref-journal36}               & 26.2505 ± 8.5938  & 4.0128 ± 1.2187 & {\bfseries 29.3884 ± 10.34}   & 251.7723 ± 162.9558 \\
			Dichotomy               & 26.0788 ± 9.3296  & 4.4876 ± 1.3338 & 42.7569 ± 10.2974 & 445.1974 ± 401.6085 \\
			Dichotomy+filter        & 40.7416 ± 6.6555  & 4.9063 ± 0.87046  & 60.8422 ± 11.5428           & 610.7531 ± 398.9431       \\
			\multicolumn{1}{l}{Dichotomy Tuna} & {\bfseries 25.9274 ± 9.0373}  & {\bfseries 4.1882 ± 1.1575}   & {\bfseries 39.4064 ± 11.7832}   & {\bfseries 372.0437 ± 169.7588}    \\
			\hline  
		\end{tabular}%
	}
	\caption{This table provides the final results of selected methods for LOLv2-synthetic considering the no-reference metrics.}
	\label{Table6}
\end{table}

\begin{figure}[h]
	\centering
	\includegraphics[width=15.0cm]{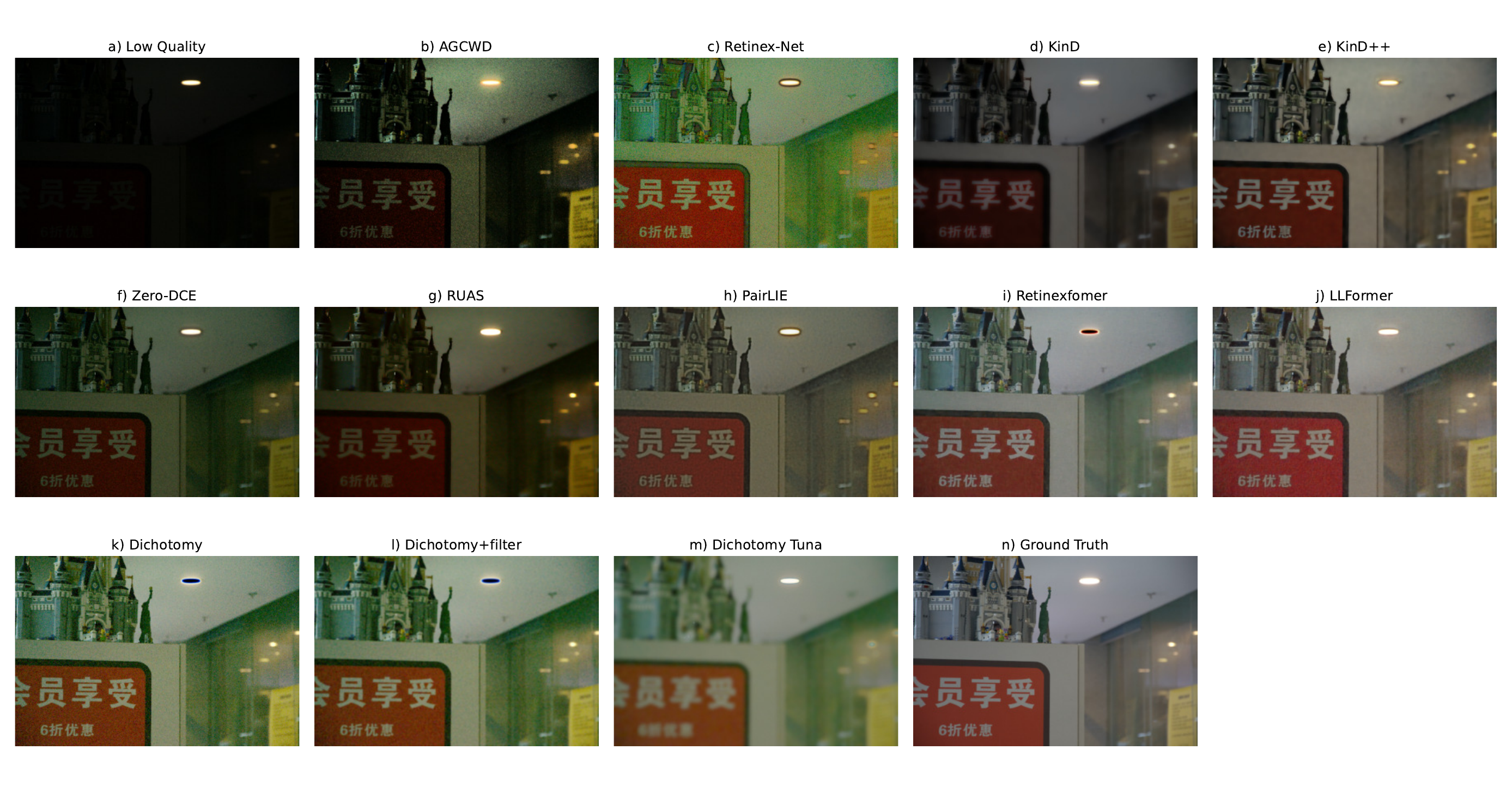}
	\caption{This collage shows the results of twelve low-light image enhancement algorithms tested on one image of the LOLv2-real database studied in this work. We can observe the problem of inverting the pixel values of Dichotomy+filter for overexposed regions and how Dichotomy Tuna solves the problem, although filtering and color issues remain.
	\label{fig5}}
\end{figure}  

\begin{figure}[h]
	\centering
	\includegraphics[width=15.0cm]{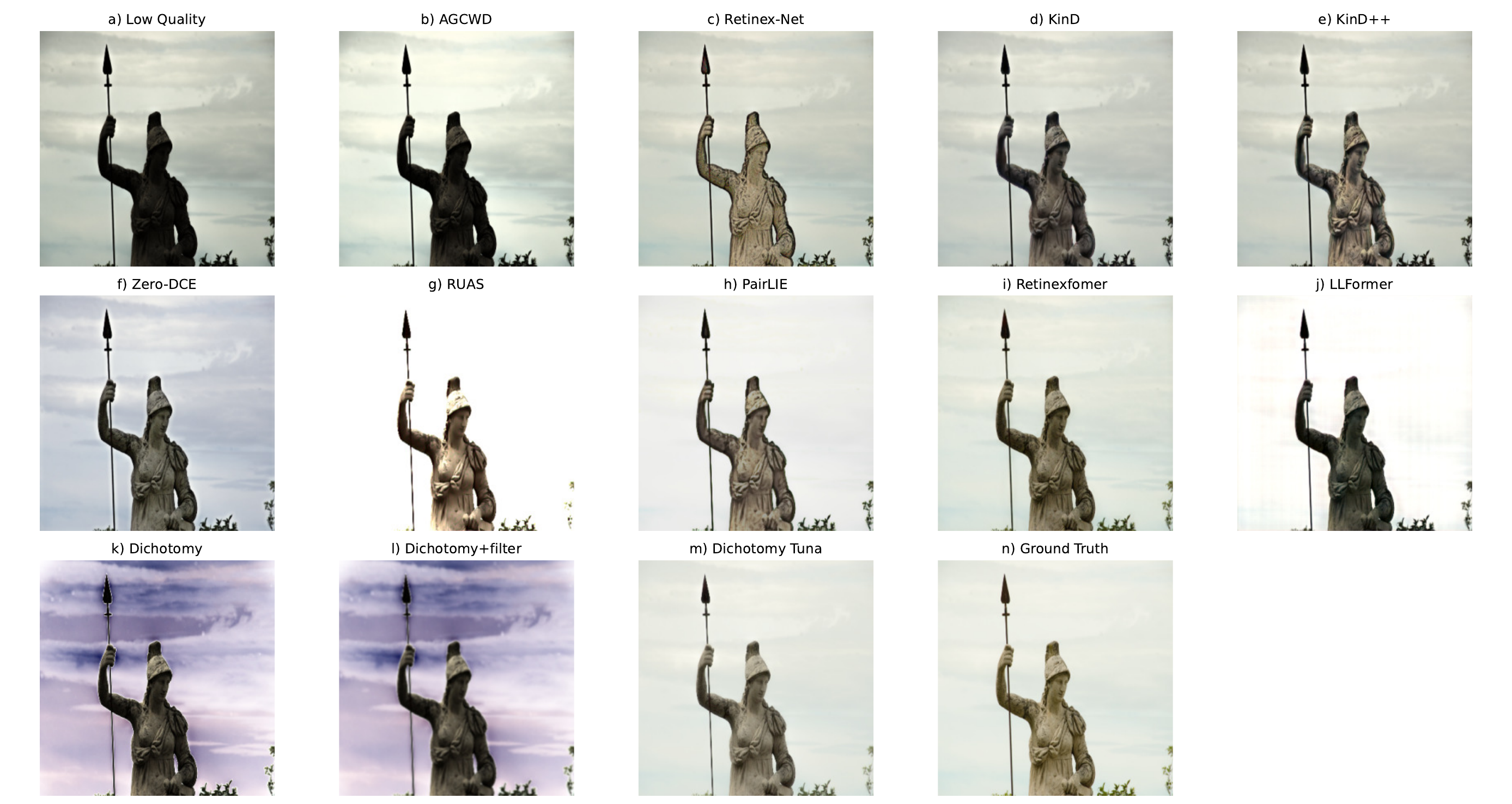}
	\caption{This collage shows the results of twelve low-light image enhancement algorithms tested on one image of the LOLv2-synthetic database studied in this work. Surprisingly, Dichotomy Tuna reproduces ground truth with simple analytical-heuristic reasoning that contrasts state-of-the-art models based on abstruse reasoning. \label{fig6}}
\end{figure}  

\begin{figure}[h]
	\centering
	\includegraphics[width=15.0cm]{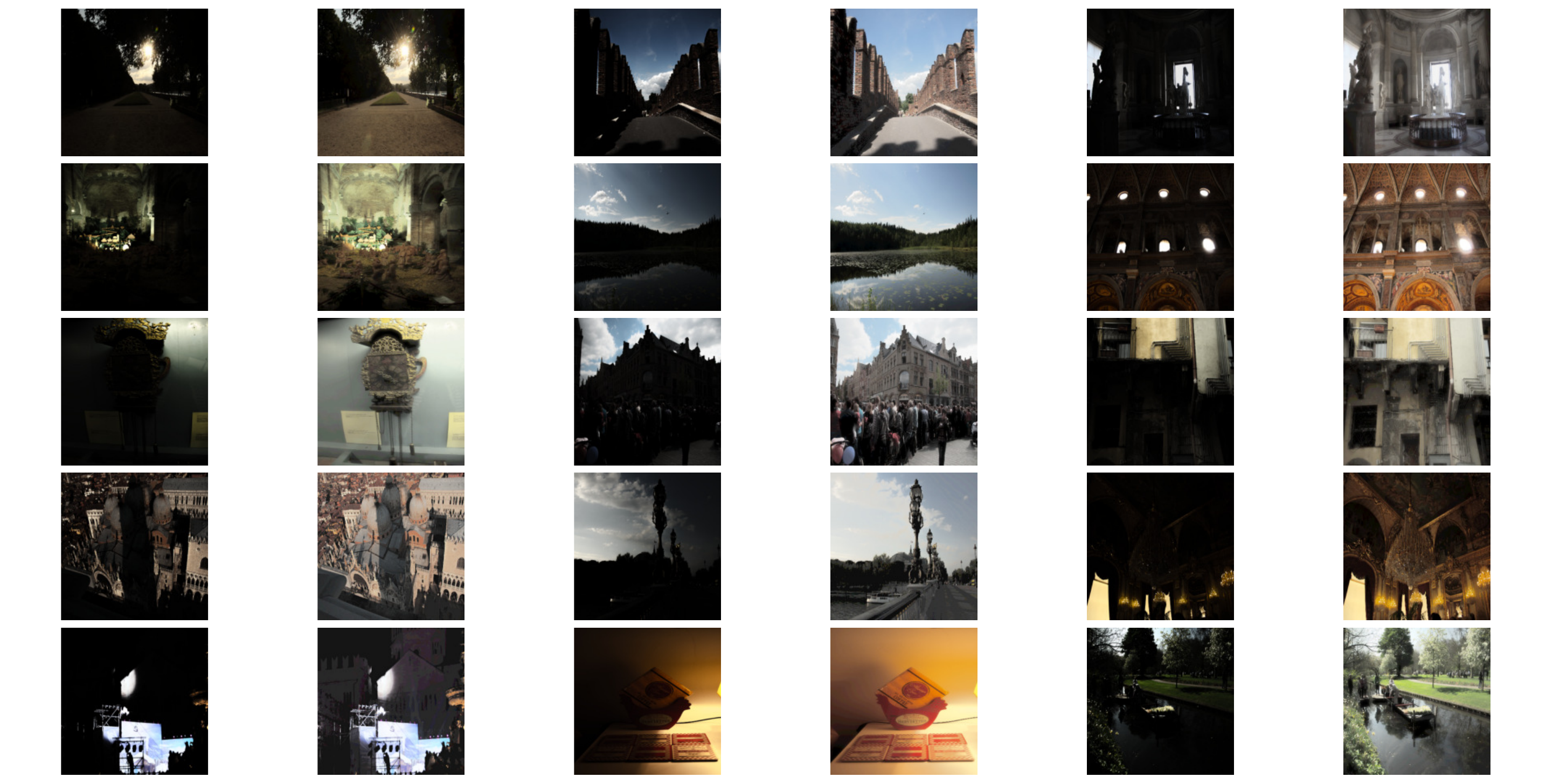}
	\caption{This collage shows the results of Dichotomy Tuna applied to 15 images selected from LOLv2-synthetic to illustrate the algorithm output. 
	\label{fig7}}
\end{figure}

\begin{figure}[h]
	\centering
	\includegraphics[width=15.0cm]{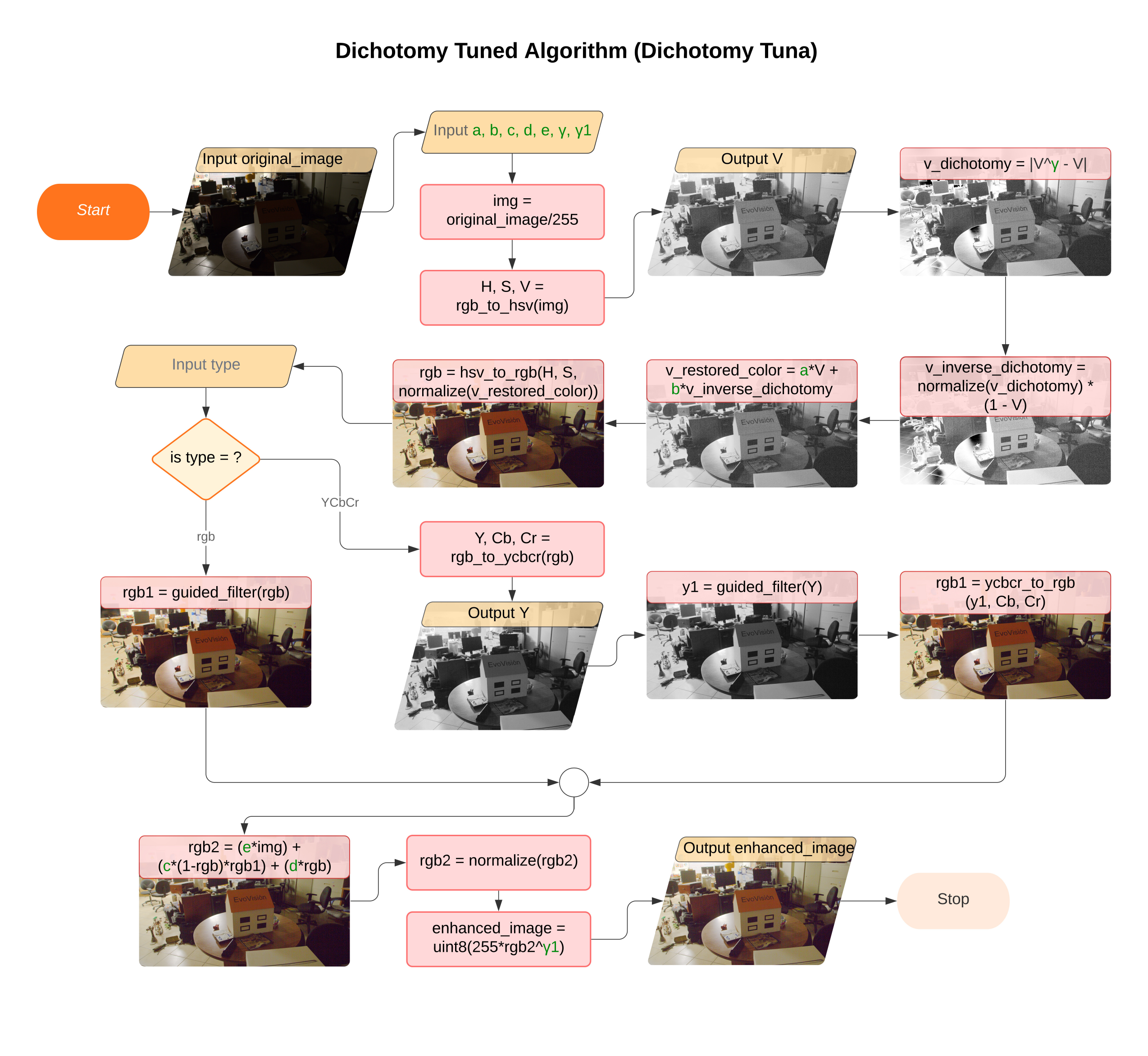}
	\caption{This flowchart shows the results of Dichotomy Tuna applied to an input image to sketch the step-by-step algorithm.
		\label{fig8}}
\end{figure}

In general, most articles show some images to get an idea of the quality of results from a visual viewpoint. Figures \ref{fig5} and \ref{fig6} report results for LOLv2 after selecting one image from each dataset and making a visual comparison with the selected twelve algorithms. Figure \ref{fig5} shows that our proposed methods based on the Dichotomy model greatly enhance all details in the image. Note that we did not tune the filter's parameters, and it is possible to improve the quality of the result further. Nevertheless, it is competitive against most recent deep learning approaches with a radical, straightforward approach that provides clear and understandable explanations. Figure \ref{fig6} provides a clear example of why Dichotomy Tuna is the best method of all proposals while improving the deficiencies of our previous approach. However, it is fair to say that we developed the original Dichotomy model to enhance visual information for many other computer vision and image processing tasks. The primary purpose of low-light image enhancement is to obtain an image whose characteristics resemble a well-taken picture under the best illumination circumstances. Our proposed strategy shows the best approach to studying this problem. Figure \ref{fig7} shows fifteen selected images out of the 100 that LOLv2-synthetic contains to illustrate the final results of Dichotomy Tuna. Figure \ref{fig8} provides the step-by-step flowchart with an example of the EvoVisión house.

\section{Discussion and Conclusion}

One could acquire an image under suboptimal lighting conditions for many reasons, including environmental or technical constraints that result in insufficient or uneven illumination. Finding a cognoscible and straightforward solution to low-light image enhancement could open new possibilities to numerous application areas like visual surveillance, autonomous driving, and computational photography. The proposed technique opens an avenue for swarm and evolutionary computation where metaheuristics could shine in challenging computer vision tasks where poor illumination occurs. The endeavor consists of choosing an appropriate algorithm, proposing new evaluation metrics, and testing datasets while considering computational performance and algorithmic complexity. For people interested in classical optimization or genetic and deep learning, we suggest blending the analytical model with your favorite methodology. Tuning filtering is necessary, but we choose to apply such methods without much care since we want to highlight the value of the analytical model. 

Other open issues and future research directions include the following. 1) evaluating and contrasting the generalization capability among different datasets and computer vision tasks. 2) Removing noise and artifacts produced by physical and computational image processing. 3) Correcting uneven illumination for real-world tasks. 4) Distinguishing and integrating semantic visual information related to other physical variables like color. 5) Extending the benchmarks to challenging low-light video enhancement where real-time is a must. 

As a final thought, researchers understand that the low-light image enhancement problem is sufficiently complex. This article illustrates how to solve it with metaheuristic and analytical reasoning. 

\section{Patents}
Gustavo Olague and Axel Martinez. Sistema para el tratamiento de imágenes sin información visible al ojo humano. Instituto Mexicano de la Propiedad Intelectual. Solicitud de patente MX/a/2024/004585.

\section{Acknowledgments}
The following project support this research: Project titled "MOVIE: Modelado de la visión y la evolución" CICESE-634135.


\begin{thebibliography}{00}


\bibitem{Olague2016}
  Olague, G.
  \textit{Evolutionary Computer Vision: The First Footprints},
  Springer,
  2016.

\bibitem{Li2022}
Li, C., Guo, C., Han, L., Jiang, J., Cheng, M.M., Gu, J., and Loy, C.C. Low-light Image and Video Enhancement using Deep Learning: A Survey. {\em IEEE Transactions on Pattern Analysis and Machine Intelligence}, Vol. 44, No. 12, pp. 9396-9416, 2022. https://doi.ieeecomputersociety.org/10.1109/TPAMI.2021.3126387

\bibitem{Guo2023}
Guo, J., Ma, J., García-Fernández, A.F., Zhang, Y., and Liang, H. A Survey on Image Enhancement for Low-light Images. {\em Heliyon}. Vol. 9, Issue 4, April 2013. https://doi.org/10.1016/j.heliyon.2023.e14558

\bibitem{Ye2024}
Ye, J., Qiu, C., and Zhang, Z. A Survey on Learning-based Low-light Image and Video Enhancement. {\em Displays}. Vol. 81, January 2024. https://doi.org/10.1016/j.displa.2023.102614

\bibitem[Huang(2013)]{ref-journal5}
Huang, S. C., Cheng, F. C., and Chiu, Y. S. Efficient Contrast Enhancement using Adaptive Gamma Correction with Weighting Distribution. {\em IEEE Transactions on Image Processing}. Vol. 22, No. 3, March 2013. https://doi.org/10.1109/TIP.2012.2226047

\bibitem[Wang(2013)]{ref-journal14}
Wang, S., Zheng, J., Hu, H. M., and Li, B. Naturalness Preserved Enhancement Algorithm for Non-Uniform Illumination Images. {\em IEEE Transactions on Image Processing}, Vol. 22, No. 9, pp. 3538-3548, 2013. https://doi.org/10.1109/TIP.2013.2261309.

\bibitem[Fu(2016)]{ref-journal15}
Fu, X., Zeng, D., Huang, Y., Zhang, X. P., and Ding, X. A Weighted Variational Model for Simultaneous Reflectance and Illumination Estimation. {\em IEEE Conference on Computer Vision and Pattern Recognition}, pp. 2782-2790, 2016. https://doi.org/10.1109/CVPR.2016.304.

\bibitem[Fu(2016)]{ref-journal16}
Fu, X., Zeng, D., Huang, Y., Liao, Y., Ding, X., and Paisley, J. A fusion-based enhancing method for weakly illuminated images. {\em Signal Processing}, Vol. 129, pp. 82-96, 2016. https://doi.org/10.1016/j.sigpro.2016.05.031.

\bibitem[Guo(2016)]{ref-journal17}
Guo, X., Li, Y., and Ling, H. LIME: Low-Light Image Enhancement via Illumination Map Estimation. {\em IEEE Transactions on Image Processing}, Vol. 26, No. 2, pp. 982-993, 2016. https://doi.org/10.1109/TIP.2016.2639450. 

\bibitem[Hao(2020)]{ref-journal20}
Hao, S., Han, X., Guo, Y., Xu, X., and Wang, M. Low-Light Image Enhancement With Semi-Decoupled Decomposition. {\em IEEE Transactions on Multimedia}, Vol. 22, No. 12, pp. 3025-3038, 2020. https://doi.org/10.1109/TMM.2020.2969790. 

\bibitem[Wei(2018)]{ref-journal18}
Wei, C., Wang, W., Yang, W., and Liu, J. Deep retinex decomposition for low-light enhancement. {\em British Machine Vision Conference}, pages 1-12, 2018.

\bibitem[Xu(2020)]{ref-journal19}
Xu, K., Yang, X., Yin, B., and Lau, R. W. H. Learning to Restore Low-Light Images via Decomposition-and-Enhancement. {\em IEEE/CVF Conference on Computer Vision and Pattern Recognition}, pp. 2278-2287, 2020. https://doi.org/10.1109/CVPR42600.2020.00235.

\bibitem[Wang(2004)]{ref-journalquality}
Wang, Z., Bovik, A. C. Sheikh, H. R., and Simoncelli, E. P. Image quality assessment: from error visibility to structural similarity. {\em IEEE Transactions on Image Processing}, Vol. 13, No. 4, pp. 600-612, 2004. https://doi.org/10.1109/TIP.2003.819861.

\bibitem[Moran(2020)]{ref-journal21}
Moran, S., Marza, P., McDonagh, S., Parisot, S., and Slabaugh, G. Deeplpf: Deep local parametric filters for image enhancement. {\em IEEE/CVF Conference on Computer Vision and Pattern Recognition}, pp. 12823-12832, 2020. https://doi.org/10.1109/CVPR42600.2020.01284.

\bibitem[Guo(2020)]{ref-journalZdce}
Guo, C., Li, C., Guo, J., Loy, C.C., Hou, J., Kwong, S., and Cong, R. Zero-Referenced Deep Curve Estimation for Low-Light Image Enhancement. {\em IEEE/CVF Conference on Computer Vision and Pattern Recognition}, pp. 1780-1789, 2020. https://doi.org/10.1109/CVPR42600.2020.00185.

\bibitem[Chen(2021)]{ref-journal22}
Chen, H., Wang, Y., Guo, T., Xu, C., Deng, Y., Liu, Z., Ma, S., Xu, C., and Gao, W. Pre-trained image processing transformer. {\em IEEE/CVF Conference on Computer Vision and Pattern Recognition}, pp. 12294-12305, 2021. https://doi.org/10.1109/CVPR46437.2021.01212.

\bibitem[Yang(2021)]{ref-journal23}
Yang, W., Wang, W., Huang, H., Wang, S., and Liu, J. Sparse gradient regularized deep retinex network for robust low-light image enhancement. {\em IEEE Transactions on Image Processing}, Vol. 30, pp. 2072-2086, 2021. https://doi.org/10.1109/TIP.2021.3050850.

\bibitem[Jiang(2021)]{ref-journal24}
Jiang, Y., Gong, X., Liu, D., Cheng, Y., Fang, C., Shen, X., and Yang, J. Enlightengan: Deep light enhancement without paired supervision. {\em IEEE Transactions on Image Processing}, Vol. 30, pp. 2340-2349, 2021. https://doi.org/10.1109/TIP.2021.3051462.

\bibitem[Liu(2021)]{ref-journal25}
Liu, R., Ma, L., Zhang, X., Fan, X., and Luo, Z. Retinex-inspired Unrolling with Cooperative Prior Architecture Search for Low-light Image Enhancement. {\em IEEE/CVF Conference on Computer Vision and Pattern Recognition}, pp. 10556-10565, 2021. https://doi.org/10.1109/CVPR46437.2021.01042.

\bibitem[Wang(2022)]{ref-journal26}
Wang, Z., Cun, X., Bao, J., Zhou, W., Liu, J., and Li, H. A general u-shaped transformerfor image restoration. {\em IEEE/CVF Conference on Computer Vision and Pattern Recognition}, pp. 17662-17672, 2022. https://doi.org/10.1109/CVPR52688.2022.01716.

\bibitem[Ma(2022)]{ref-journal27}
Ma, L., Ma, T., Liu, R., Fan, X., and Luo, Z. Toward fast, flexible, and robust low-light imageenhancement. {\em IEEE/CVF Conference on Computer Vision and Pattern Recognition}, pp. 5627-5636, 2022. https://doi.org/10.1109/CVPR52688.2022.00555.

\bibitem[Xu(2022)]{ref-journal28}
Xu, X., Wang, R., Fu, C.W., and Jia, J. SNR-aware low-light image enhancement. {\em IEEE/CVF Conference on Computer Vision and Pattern Recognition}, pp. 17714-17724, 2022. https://doi.org/10.1109/CVPR52688.2022.01719.

\bibitem[Wu(2022)]{ref-journal29}
Wu, W., Weng, J., Zhang, P., Wang, X., Yang, W.H., and Jiang, J. Uretinex-net: Retinex-based deep unfolding network for low-light image enhancement. {\em IEEE/CVF Conference on Computer Vision and Pattern Recognition}, pp. 5901-5910, 2022. https://doi.org/10.1109/CVPR52688.2022.00581.

\bibitem[Lei(2022)]{ref-journal30}
Lei, X. Fei, Z., Zhou, W., Zhou, H., and Fei, M. Low-light image enhancement using the cell vibration model. {\em IEEE Transactions on Multimedia}, Vol. 25, pp. 4439-4454, 2022. https://doi.org/10.1109/TMM.2022.3175634.

\bibitem[Yang(2021)]{ref-journal31}
Yang, W., Wang, S., Fang, Y., Wang, Y., and Liu, J. Band representation-based semi-supervised low-light image enhancement: Bridging the gap between signal fidelity and perceptual quality. {\em IEEE Transactions on Image Processing}, Vol. 30, pp. 3461-3473, 2021. https://doi.org/10.1109/TIP.2021.3062184.

\bibitem[Zhang(2019)]{ref-journal32}
Zhang, Y., Zhang, J., and Guo, X. Kindling the darkness: A practical low-light image enhancer. {\em ACM International Conference on Multimedia}, pp. 1632-1640, 2019. https://doi.org/10.1145/3343031.3350926.

\bibitem[Zhang(2021)]{ref-journalknd++}
Zhang, Y., Guo, X., Ma, J., Liu, W., and Zhang, J. Beyond Brightening Low-light Images. {\em International Journal of Computer Vision}, Vol. 129, pages 1013-1037, 2021. https://link.springer.com/article/10.1007/s11263-020-01407-x.


\bibitem[Fu(2023)]{ref-journal33}
Fu, Z., Yang, Y., Tu, X., Huang, Y., Ding, X., and Ma, K.K. Learning a Simple Low-light Image Enhencer from a Paired Low-light Instances. {\em IEEE/CVF Conference on Computer Vision and Pattern Recognition}, pp. 22252-22261, 2023. https://doi.org/10.1109/CVPR52729.2023.02131.

\bibitem[Cai(2023)]{ref-journal35}
Cai, Y., Bian, H., Lin, J., Wang, H., Timofte, R., and Zhang, Y. Retinexformer: One-stage Retinex-based Transformer for Low-light Image Enhancement. {\em IEEE/CVF International Conference on Computer Vision}, pp. 12504-12513, 2023.  https://doi.ieeecomputersociety.org/10.1109/ICCV51070.2023.01149.

\bibitem[Cai(2023)]{ref-journal36}
Wang, T., Zhang, K., Shen, T., Luo, W., Stenger, B., and Lu, T. Ultra-High-Definition Low-Light Image Enhancement: A Benchmark and Transformer-Based Method. {\em Proceedings of the 37th AAAI Conference on Artificial Intelligence}, AAAI Press, Vol. 37, Issue 3, 2023. https://doi.org/10.1609/aaai.v37i3.25364.

\bibitem[Wang(2023)]{ref-journal37}
Wang, Y., Liu, Z., Liu J., Xu, S.,and Liu, S. Low-light Image Enhancement with Illumination-aware Gamma Correction and Complete Image Modelling Network. {\em IEEE/CVF Conference on Computer Vision and Pattern Recognition}, pp. 13128-13137, 2023. https://doi.org/10.1109/ICCV51070.2023.01207.

\bibitem[Martinez(2024)]{unpublished2024}
Martinez, A., Olague, G., and Hernandez, E. Modeling Image Tone Dichotomy with the Power Function. Unpublished 2024.

\bibitem[Holland(1992)]{inbook2}
Holland, J.H., Adaptation in Natural and Artificial Systems. 211 pages, MIT Press (first appear in 1975), 1992.

\bibitem[Deb(1995)]{ref-journal34}
Deb, K., and Agrawal, R.B. Simulated Binary Crossover for Continuous Search Space. Complex Systems., Vol. 9, Issue 2, pp. 115-148, 1995. https://content.wolfram.com/sites/13/2018/02/09-2-2.pdf.





\end{thebibliography}
\end{document}